\def\year{2021}\relax
\documentclass[letterpaper]{article} 
\usepackage{aaai21}  
\usepackage{times}  
\usepackage{helvet} 
\usepackage{courier}  
\usepackage[hyphens]{url}  
\usepackage{graphicx} 
\usepackage{natbib}  
\usepackage{caption} 
\usepackage{amsmath}
\usepackage{subcaption}
\title{Stabilizing Transformer-Based Action Sequence Generation  \\ For Q-Learning }

\author{

    Gideon Stein \textsuperscript{\rm 1}, \\
    Andrey Filchenkov \textsuperscript{\rm 2},
    Arip Asadulaev \textsuperscript{\rm 3} \\

}

\affiliations{

    ITMO University

    St. Petersburg, 198215 \\
     \textsuperscript{\rm 1}gideon@steinml.de,  \textsuperscript{\rm 2}afilchenkov@itmo.ru,  \textsuperscript{\rm 3}aripasadulaev@itmo.ru

}

\begin{document}

\maketitle

\begin{abstract}
Since the publication of the original Transformer architecture \cite{vaswani2017attention}, Transformers revolutionized the field of Natural Language Processing. This, mainly due to their ability to understand timely dependencies better than competing RNN-based architectures. Surprisingly, this architecture change does not affect the field of Reinforcement Learning (RL), even though RNNs are quite popular in RL, and time dependencies are very common in RL. Recently, \cite{parisotto2019stabilizing} conducted the first promising research of Transformers in RL. To support the findings of this work, this paper seeks to provide an additional example of a Transformer-based RL method. Specifically, the goal is a simple Transformer-based Deep Q-Learning method that is stable over several environments. Due to the unstable nature of Transformers and RL, an extensive method search was conducted to arrive at a final method that leverages developments around Transformers as well as Q-learning. The proposed method can match the performance of classic Q-learning on control environments while showing potential on some selected Atari benchmarks. Furthermore, it was critically evaluated to give additional insights into the relation between Transformers and RL.
\end{abstract}


     \noindent Transformer architectures revolutionized the field of Natural Language Processing (NLP). The classic Transformer \cite{vaswani2017attention} and its successors such as \cite{devlin2018bert} or \cite{radford2019language} outperform traditionally used RNN-based architectures on the majority of tasks in NLP. Their superior performance can be mostly attributed to their ability to understand timely dependencies and notably long-term dependencies better than RNN-based methods. Timely dependencies are not only interesting for NLP tasks. In reinforcement learning, such an ability is useful to perform in environments that are only partially observable. RNN-based methods are traditionally deployed to such environments. Furthermore, there are multiple successful examples of applications of the Attention mechanism (the core functionality of the Transformer) to RL \cite{iqbal2019actor}, \cite{oh2016control} and \cite{manchin2019reinforcement}. Due to these facts, the deployment of Transformer-based architectures to RL is a promising research direction. While there were several studies on Transformer-based methods in RL, many of them such as   ~\cite{upadhyay2019transformer} or ~\cite{mishra2017simple} reported random performance for their Transformer-based approaches. Even on simple MDP or Multi-armed Bandit problems. Contrary to that, the first major success with Transformer-based RL methods was recently reported by ~\cite{parisotto2019stabilizing}. So while it is possible to use Transformer-based architectures in RL, it seems to be a nontrivial task. 
In RL, the information signal is affected by past decisions. This creates dependencies and makes optimization harder than training on a fixed dataset. Additionally, Transformer-architectures are quite hard to optimize which was already stated in \cite{vaswani2017attention}. As an example, they are strongly dependent on a specific learning rate schedule to be optimized. Together, these two conditions make the optimization of Transformer-based architectures in RL challenging. 

To support the results of ~\cite{parisotto2019stabilizing} and to further evaluate the possibility of Transformer-based models in RL, this paper seeks to create a new Transformer-based RL method that contrary to ~\cite{parisotto2019stabilizing} features a model that is based on the original Transformer from \cite{vaswani2017attention} instead of the Transformer-XL \cite{dai2019transformer}. Furthermore, this work will alternatively use the Transformer-based model as a value function for a Deep Q-learning agent. While the approach of ~\cite{parisotto2019stabilizing} showed strong performance, its optimization method as well as their Transformer model are quite advanced. On contrary, the approach in this paper uses a well-known optimization method (Q-learning) as well as a simple Transformer version to add to a clear understanding of the interrelations between RL and Transformer-based models. Furthermore, it is hoped that this choice will make it easier to retrace results that are reported in this paper and encourage additional research. In summary, the following ideas are hoped to be supported: 
\begin{itemize}
\item With a couple of alterations, Transformers can generally perform in RL.
\item Transformer-based Deep Q-networks (TBQN) can perform.
\item Transformers can outperform RNN based models in RL.
\end{itemize}

\section{Background}

\subsection{Deep Q-learning}
The goal of Q-learning is to find a function that correctly maps state (s) action (a) pairs to their corresponding value for an agent that interacts with an environment.
The simplest form of Q-learning is defined as updating a Q-function in the following manner:

 \begin{equation} \label{eq:1}
\begin{split}
   td=  & R_{t+1} +  \gamma \max_{a} Q(s_{t+1}, a) -  Q(s_{t}, a_{t} )\\
  &Q(s_{t},a_{t}) \leftarrow  Q(s_{t},a_{t})  + \alpha *  td \\
 \end{split}
 \end{equation}

where $R$ is a reward, $\gamma$ is a discount factor and $Q$ represents the value of any pair (s,a). By updating Q-values after rewards, the greedy policy as-well as the value function change frequently. Under the condition that all states are explored sufficiently, these updates are guaranteed to converge to a Q-function that correctly represents the environment. Based on this Q-function, a policy will be formed by greedily sampling the action with the highest Q-value at every step. Traditionally, the Q-function was implemented as a table. However, it is possible to approximate it with a neural network. This method is known as Deep Q-learning. When using a function approximator for the Q-function, the definition of an update changes since it is only possible to update weights of the network and not specific Q-values directly. An update for a Deep Q-network (DQN) is therefore defined as:

\begin{equation}  \label{eq:2} 
\begin{split}
  &target =  R(s,a,s') + \gamma \max_{a'} q_{k}(s',a')\\
  &td = \left[(q_{\theta}(s, a) - target)^2 \right] \\
 &\theta_{k+1} \leftarrow  - \alpha * \nabla_{\theta} \mathbf{E}_{s'~ P(s' | s,a)} td \bigg|_{\theta = \theta_{k}} \\
 \end{split}
 \end{equation}

where $\theta $ represents the network weights and $s'$ and $a'$ are the state and action in the timestep t+1. When a network is updated according to Formula \ref{eq:2}, an issue arises. Data that is acquired by an agent interacting with an environment is quite different from a fixed dataset that is normally used to train neural networks. Today, Replay Buffers, a method to save experience and reuse it during model training, and target networks, a method to make the target of the update more stable, are used to counter these issues. By applying these two methods to Deep Q-learning, \cite{mnih2013playing}  opened the field of Deep Reinforcement Learning. Their method will be the baseline RL method for the course of this work. 

\subsection{The Transformer Architecture}
\begin{figure}
\centering
\includegraphics[width=0.15\textwidth]{
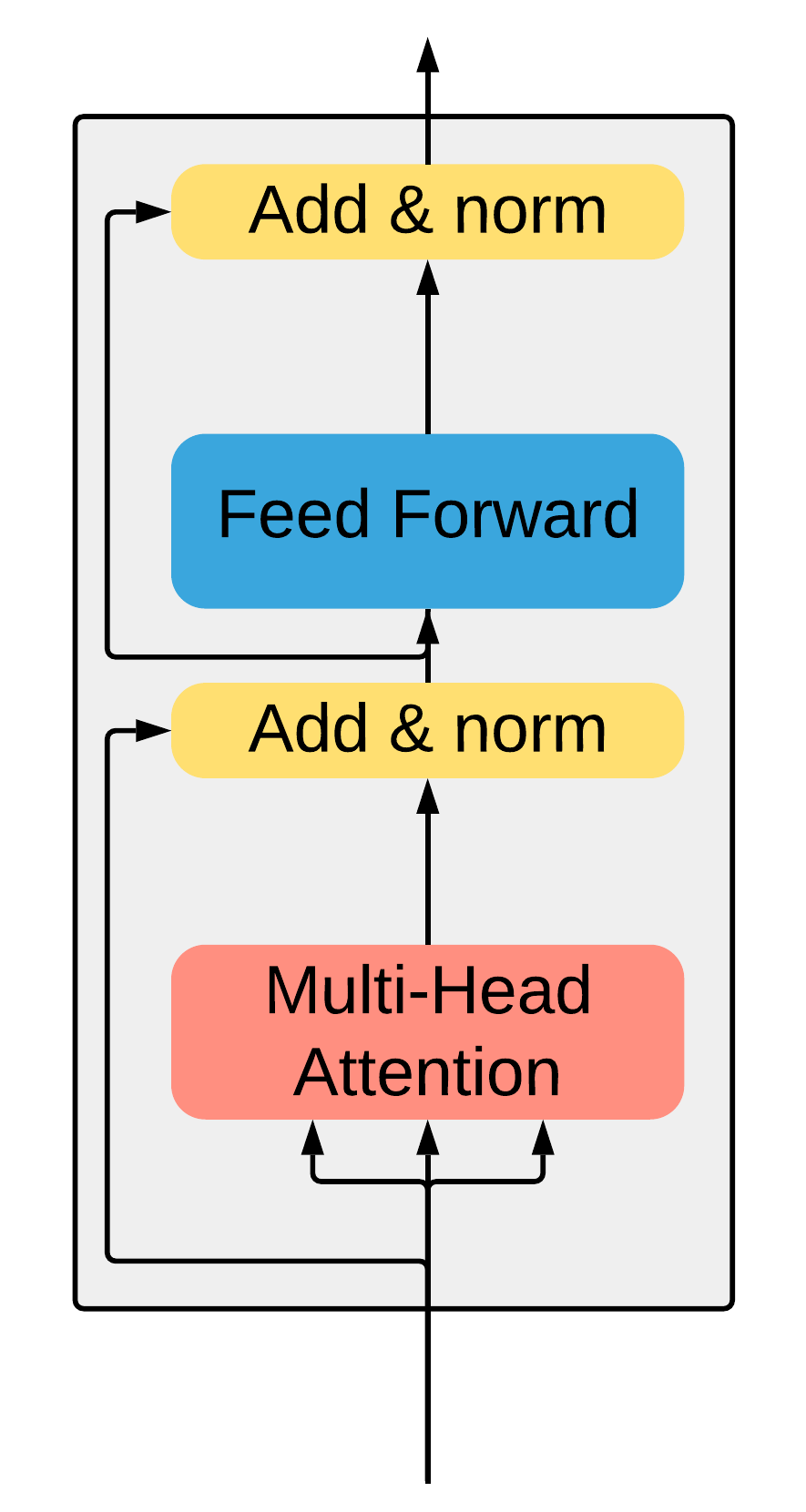}
\caption{The standard Transformer Encoder layer}
\label{fig:encoder_l}
\end{figure}

 The Transformer model is a sequence to sequence architecture (seq2seq) which was initially developed to perform translation tasks in NLP, and which relies heavily on the Attention mechanism. A seq2seq structure is defined as a model that takes in a sequence of signals and returns a sequence of outputs. Also, the Transformer is an Encoder-Decoder structure that splits into two distinct submodels.  An Encoder, that transforms an input sequence into an encoded representation and a Decoder that generates, based on the encoded representation, a new sequence as an output. Since the proposed architecture is based on the Encoder of the classic Transformer \cite{vaswani2017attention}, its structure will be discussed further. 
 The Transformer Encoder takes in some word tokens and transforms them into the same number of encoded representations. To do this efficiently, the Transformer stacks several identical blocks on top of each other. These blocks are called Encoder layers. Additionally, the Encoder features an Embedding layer and a positional encoding of the input sequences which both are added before the first layer. A single Encoder layer is constructed out of two main components. An Attention block and a feed-forward network. Additionally, residual connections and normalization are added. Fig. \ref{fig:encoder_l} shows the structure of the Encoder layer. Note that the input and the output of the Encoder have the same dimension. This makes layer stacking possible. The computation that takes place in a single Encoder layer is defined as:

 \begin{align} \label{eq:3}
\begin{split}
  out_1 &=  Norm(Attention(X)  + X )\\
   out_2 &= Norm(FF(out_1) + out_1) \\
 \end{split}
 \end{align} 
 
where X represents an input tensor with the shape (batch size, input sequence length, model dimension). Typically, Dropout is deployed after the Attention block and after the feed-forward block. 

\subsection{The Attention mechanism}

 Attention is a mechanism that understands the importance of specific inputs for other inputs and combines these into a new vector that includes this information. This mechanism does not rely on a hidden representation that includes all past information but attends directly to the full inputs. This helps Attention to perform better than RNN-based approaches in many cases,  especially when long-term dependencies are present and relevant. Based on Attention, Multi-Head Attention is performed by multiple Attention operations in parallel on sub-parts of the inputs. This allows attending multiple sub-areas of inputs at once. The Transformer features the use of Scaled Dot Product Attention as well as Multi-Head Attention to understand dependencies. Scaled dot product Attention is defined as the operation on three inputs. Keys (K), Queries (Q), and Values (V): 
\begin{align} \label{eq:attn_1}
\begin{split}
 out_1 &=  QK^T\\
 out_2 &=  out_1 / \sqrt{dim_{key}} \\
 out_3 &=  softmax(out_2)V \\
 \end{split}
 \end{align} 
 
Where Q, K, V are input matrices, and $dim_{key}$ is the last dimension of K. Intuitively, this can be understood as a way to scale and add the content of V by a factor that is a combination of Q and V. Through this channel, V attends to the information that is included in Q and K and is altered accordingly. To perform Multi-Head Attention, the initial input vector is simply split. When performing Multi-Head Attention in the Encoder, the embedded input sequence represents K, Q, and V. This specific form of Attention is called Self-Attention, since the input sequence attends to itself. It is a key component that allows the Encoder to encode the input sequence efficiently.


\section{Transformers for Q-learning}
\subsection{Transformer-based Q-Networks}

\begin{figure}
\centering
\includegraphics[width=0.2\textwidth]{
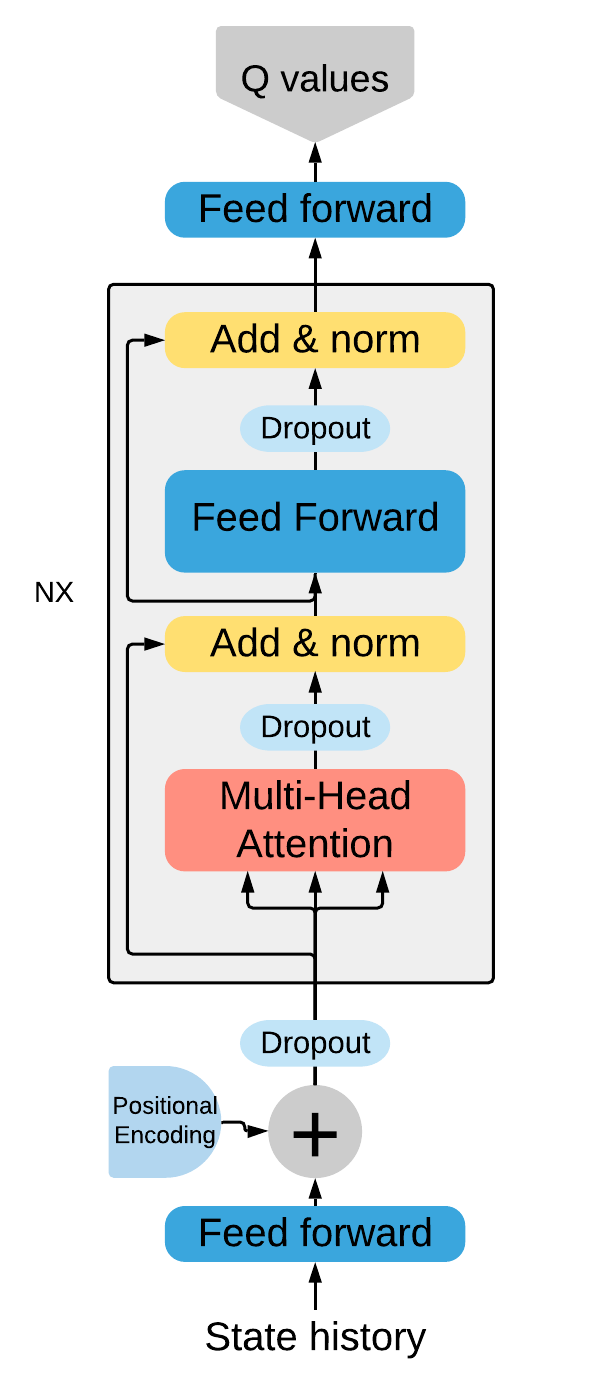}
\caption{The proposed Transformer-based Q-network}
\label{fig:trans_base}
\end{figure}

This paper proposes to use an altered version of the Transformer Encoder as a Q-network for a Q-learning agent. However, the original structure has to be altered slightly to be usable.
To map to Q-values at the end of the model, the output of the Encoder has to be mapped to the Q-value dimension which is achieved by adding a fully connected layer after the last Encoder layer. Also, the embedding layer of the classic Transformer has to be replaced by a fully connected layer that maps from the state dimension to the model dimension. After these two steps, a Transformer-based Q-network (TBQN) that can map from states to Q-values is obtained. It can be examined in Figure \ref{fig:trans_base}. 

The literature \cite{parisotto2019stabilizing},~\cite{upadhyay2019transformer}, ~\cite{mishra2017simple} suggests, that a Q-learning agent using the proposed TBQN would be very hard to optimize and most likely unstable. To preemptively counter this, a method variation search space was constructed which includes three categories. Firstly, changes to the model structure itself. Secondly, the application of additional methods for DQNs and Transformers. Both of these categories represent small model or method variations that are proposed in the literature and might be able to improve the performance of TBQNs. Thirdly, a selection of possible impactful Hyperparameters is included. This search space was then filtered to find a method variation that is easier to optimize and more stable than a base Q-learning agent featuring the base TBQN.

\subsection{Transformer layer variations}
Since the original publication of the Transformer ~\cite{vaswani2017attention}, many Transformer layer variations were introduced in the literature. These structural changes are exclusively made to make the Transformer more stable during training. From this literature, several Transformer layer variations 
were selected to be tested as the core layer for TBQNs.

\subsubsection{Dropout free models (layer type 2)}
Since Transformers were initially developed for NLP, they feature the usage of Dropout layers. Typically implemented to counter overfitting, the usage of Dropout in RL is not popular. Due to this, a layer without Dropout was tested. Additionally, all layer variations are tested with and without Dropout after the final layer. This layer variation is displayed in Figure \ref{fig: architecture variations}a.

\subsubsection{Identity Map Reordering (IMR) (layer type 3)} 
A layer variation that was described in ~\cite{parisotto2019stabilizing}. It features the positional change of the normalization layer to the start of each sub-layer. Furthermore, an additional ReLU activation after every sub-layer was added to prevent two linear layers in a row. Its implementation can be observed in Figure \ref{fig: architecture variations}b.

\subsubsection{Pre layer Normalization  (layer type 4)}
\begin{figure} 
\centering
\begin{subfigure}{.23\textwidth}
  \centering
  \includegraphics[width=0.73\linewidth]{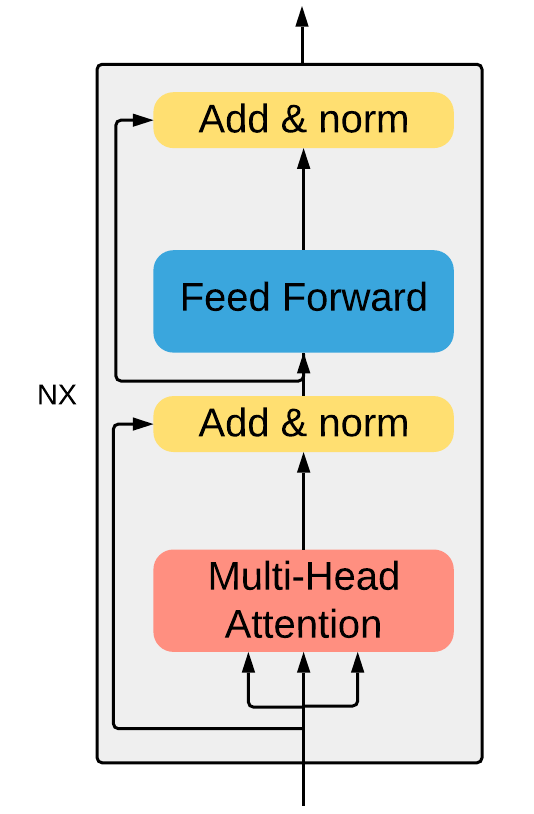}
  \caption{Layer type 2 (no dropout)}
\end{subfigure}%
\begin{subfigure}{0.23\textwidth}
  \centering
  \includegraphics[width=0.6\linewidth]{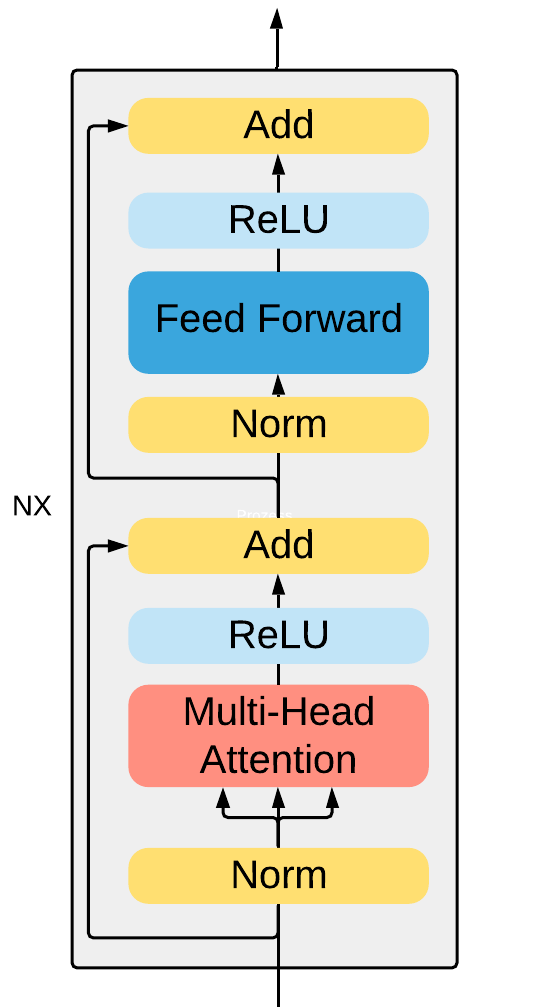}
  \caption{Layer type 3 (IMR)}
\end{subfigure}
\begin{subfigure}{.23\textwidth}
  \centering
  \includegraphics[width=0.77\linewidth]{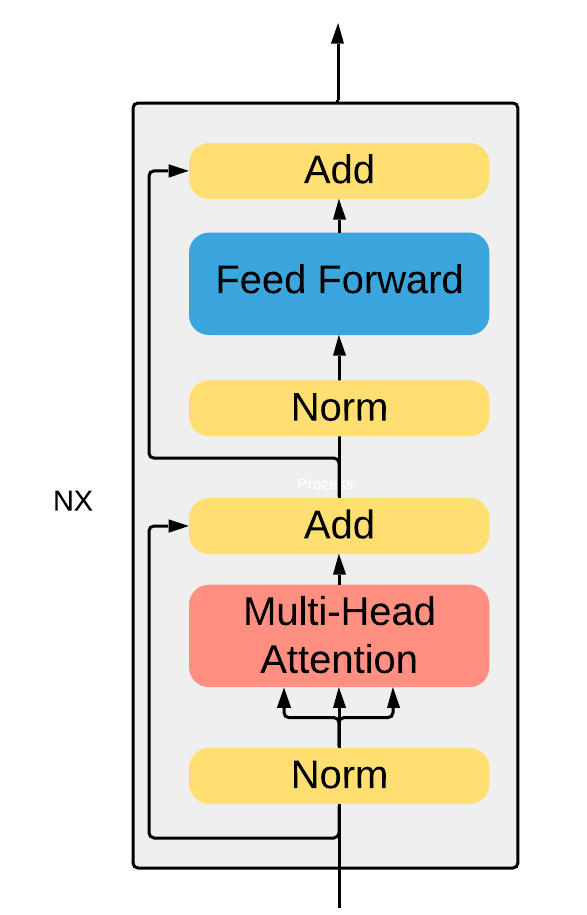}
  \caption{Layer type 4 (norm first)}
\end{subfigure}%
\begin{subfigure}{.23\textwidth}
  \centering
  \includegraphics[width=0.6\linewidth]{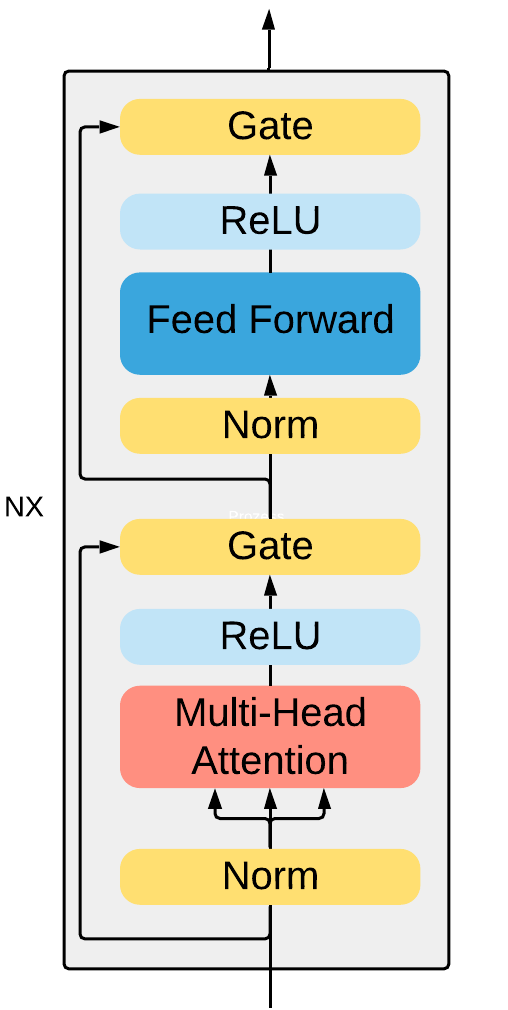}
  \caption{Layer type 5/6 (gated)}
\end{subfigure}%
\caption{The Transformer Encoder layer variations}
\label{fig: architecture variations}
\end{figure}

Very similar to IMR, this variation described in \cite{xiong2020layer} changes the position of the layer normalization to the beginning of each sub-layer. While this is identical to IMR, this variation does not feature an additional ReLU activation. Its implementation can be observed in Figure \ref{fig: architecture variations}c.

\subsubsection{Output gate connections  (layer type 5)}
Also described in ~\cite{parisotto2019stabilizing} this variation based on IMR additionally replaces the residual connection with a gated layer. While residual connections were initially implemented to improve the training of deep neural networks, they seem to make training Transformers more unstable. They are replaced with the following gate formulation, where W and b are trainable parameters:

\begin{equation} \label{eq:5}
g_l(x,y) = x + \sigma (W_l^g x -b_l^g) \odot y
\end{equation}

This variation was also already tested for Transformer-based methods in RL and it will be used as it was proposed in  \cite{parisotto2019stabilizing}. 

\subsubsection{GRU gate connections  (layer type 6)}
Finally, another variation will be tested which features the usage of a different gating mechanism based on a GRU unit. Again, this variation was introduced in ~\cite{parisotto2019stabilizing} and is based on IMR. Noteworthy is that this model variation combined with Maximum a Posteriori Policy Optimization \cite{song2019v} achieved SOTA results for DMLab-30. It remains to be seen if this is also the case for Q-learning. The mechanism is defined by Formula (6). W and U are trainable parameters.

\begin{align} \label{eq:6}
\begin{split}
 H &=  \tanh(W_l^H y + U_l^H (R \odot x))  \\
 Z &=  \sigma(W_l^Z y +  U_l^Z x -b_l^g)  \\
 R &=  \sigma(W_l^R y + U_l^R x) \\
 g_l  &(x,y) = (1-Z)\odot + Z\odot H \\
 \end{split}
\end{align}

\subsection{Additional methods and Hyperparameters}

Additionally to these layer variations, the following methods and Hyperparameters were included categorically in the search space to test their effect on the performance of a TBQN:

\begin{itemize}
\item Double Q-learning
\item Target update period
\item Target update ($\tau$) ~\cite{lillicrap2015continuous}
\item Gradient Clipping
\item Learning rate schedules
\item Depth-Scaled Initialization ~\cite{zhang2019improving}
\item Depth-Scaled Initialization of the last Layer ~\cite{zhang2019improving}
\item Number of Attention Heads
\item Initial collection steps
\item Loss function
\item Environment normalization
\item Epsilon Greedy
\item Replay Buffer size
\item Future reward discount ($\gamma$)
\item Batch size
\item Learning rate
\item Encoder type (whether or not dropout is used outside of the Encoder layers) 
\end{itemize}

\section{Experiments}

\subsection{Baseline performance}

\begin{figure} 
\centering
\begin{subfigure}{.23\textwidth}
  \centering
  \includegraphics[width=1\linewidth]{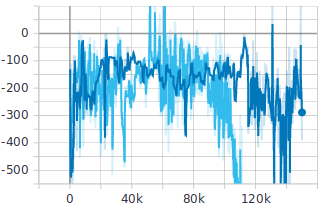}
  \caption{LunarLander-v2}
\end{subfigure}%
\begin{subfigure}{.23\textwidth}
  \centering
  \includegraphics[width=1\linewidth]{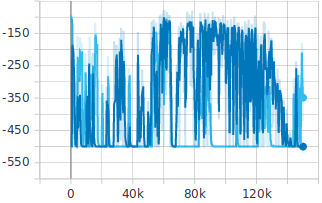}
  \caption{Acrobot-v1}
\end{subfigure}
\begin{subfigure}{.23\textwidth}
  \centering
  \includegraphics[width=1\linewidth]{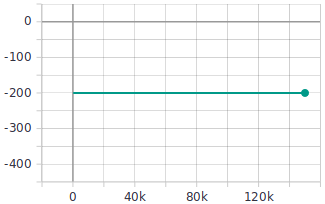}
  \caption{MountainCar-v0}
\end{subfigure}%
\begin{subfigure}{.23\textwidth}
  \centering
  \includegraphics[width=1\linewidth]{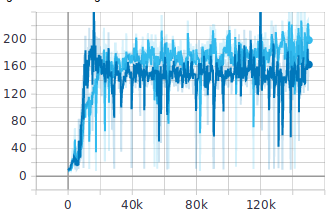}
  \caption{CartPole-v1}
\end{subfigure}%
\caption{Average return during training of a  base Q-learning agent with the proposed TBQN base in four different environments.}
\label{fig: base_performance}
\end{figure}

To motivate the method variation search and to set a base performance of TBQNs, a Q-learning agent with the proposed base TBQN and with no special additions (except a Replay Buffer and a Target Network) was evaluated. The agent was trained on four environments (MountainCar-v0, Acrobot-v1, CartPole-v1, and LunarLander-v2) for 150k steps. All these environments are implemented by OpenAI GYM ~\cite{1606.01540}. The average episode return over 10 episodes can be examined in Figure \ref{fig: base_performance} . Two training runs per environment were executed (For Acrobot-v1, only one is displayed to guarantee the visibility of the results). The agent was not able to solve any environment sufficiently, had a high fluctuation, and even diverged on some occasions (denoted by a graph ending before 150k steps). For these experiments, the following Hyperparameters were used. Initial collect steps: 1000, mean squared loss, 4 Attention Heads, epsilon greedy: 0.1, Replay Buffer length: 100000, batch size: 32, learning rate: 1e-5. The rest of the parameters were not used. It shows quite clearly, that TBQNs need additional help to perform.

\subsection{Selecting the optimal method variation}
While it would be ideal to test every possible method variation, this is unfeasible due to computational complexity. Due to that, a two-step method based on two distinct studies was constructed to find a well-performing method variation.

\subsubsection{Study one - Parameter importance}
The first study focused on narrowing down the method search space significantly. This was achieved by estimating the Mean Decrease Impurity Importance Score for all parameters in the method search space. Based on these scores, parameters with low importance were excluded entirely. Furthermore, parameters with high importance were further evaluated to select the best performing values and exclude the rest from the search space. To estimate these scores, the method search space had to be sampled and evaluated. Since grid search was infeasible, a Tree-Structured Parzen Estimator, which was firstly described in  ~\cite{bergstra2011algorithms}, was used to sample from the search space. 
 Every method search space sample was trained for 15k steps. As a final performance score, the average return of the last 10 episodes was used. To guarantee generality, the study was conducted independently in three different environments (CartPole-v1, Acrobot-v1, and LunarLander-v2) and the final importance score for every parameter was averaged between these environments. All studies were performed on a single GPU (Nvidia1080Ti). Further information can be found in Appendix B.

\subsubsection{Study two - Final selection}
After having narrowed down the method search space significantly, the remaining search space samples were evaluated further to find the model variation with the best performance. Two methods were used to determine the effect of certain parameter values on the performance of TBQNs and to select a final method variation: On one hand, the mean reward of the last ten episodes between all samples where a certain parameter value was present was calculated. On the other hand, the search space samples with the highest rewards for every environment were extracted. This was done to determine whether combinations of specific parameter values performed especially well. Again, the study was conducted in three different environments (CartPole-v1, Acrobot-v1, and LunarLander-v2) to guarantee generality. All search space samples were trained for 75K steps in the environments CartPole-v1 and AcroBot-v1 and for 150k steps in the environment LunarLander-v2. All experiments were conducted on a single Nvidia GPU(1080ti). Further information can be found in Appendix B. 

\section{Results} 
Based on the two studies, the method variation represented by Table \ref{table:3} was selected as it performed well in all environments and proved to be stable during training. 
The parameters initial collect steps, Environment normalization,  Replay Buffer size, $\tau$, double Q-learning, and the Encoder type had low importance for control environments and are not specified. The following comments should be made to accompany this selection:

\begin{table}
\centering
\begin{tabular}{ |p{3cm}||p{1.5cm}|p{2cm}|  }
 \hline
 Parameter  & Value  & Category\\
 \hline
  \hline
  Gradient Clipping & True  & Fixed\\
 Batch size & 32   & Fixed\\
 Learning rate & 1e-4  & Fixed\\
 Layer type & 3  & Fixed\\
 Custom lr schedule&"No" & Fixed \\
 Depth-Scaled Initialization &   1 & Fixed \\
Target upate period & 10+  &  Semi-fixed \\
 Num Heads & 4/2  & Semi-fixed  \\
 Epsilon Greedy & (0. - 1.)  &  Environment dependent \\
  Depth-Scaled Initialization (last layer) & (T/F) & Environment dependent\\
 Loss function &   (Huber, Squared) &  Environment dependent\\
 $\gamma$ & (.99, .95)  &  Environment dependent  \\
 \hline
\end{tabular}
\caption{Final method variation}
\label{table:3}
\label{table:Final}
\end{table}

\begin{itemize}
\item The optimal values for several parameters are environment-dependent. This means the performance of a Q-learning agent using a TBQN relies strongly on the right value selection. The optimal values however change from environment to environment.  
\item Surprisingly, IMR layers (layer type 3) perform the best while GRU-gated layers (layer type 6) were excluded early due to frequent divergence. 
\item While being very important for NLP, learning rate schedules are not required for TBQNs. It is estimated that TBQNs with layer variations do not require learning rate schedules which makes them obsolete. 
\item Depth-Scaled Initialization \cite{zhang2019improving} is beneficial. Models that were initialized with it tended to diverge less and achieved higher average rewards at the end of training.
\item Gradient Clipping is very important for TBQNs. Since the Transformer has problems with divergence in the RL setting, Gradient Clipping helps to mitigate destructive updates.
\item Several parameters are not important for model performance (Assuming no abstruse values). During the parameter search, they showed no significant impact on the performance of TBQNs. 
\end{itemize}

\subsection{Performance in control environments}

\begin{figure} [ht]
\centering
\begin{subfigure}{.233\textwidth}
  \centering
  \includegraphics[width=1\linewidth]{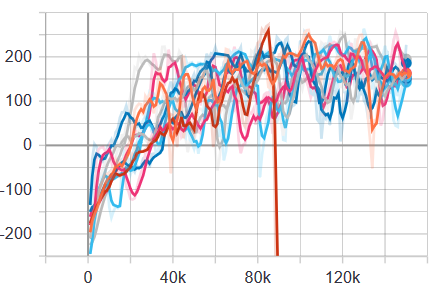}
  \caption{LunarLander-v2 (150k steps)}
\end{subfigure}%
\begin{subfigure}{.233\textwidth}
  \centering
  \includegraphics[width=1\linewidth]{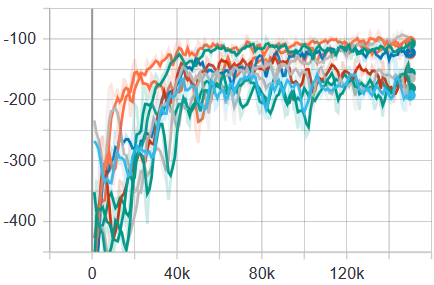}
  \caption{Acrobot-v1 (150k steps)}
\end{subfigure}
\begin{subfigure}{.233\textwidth}
  \centering
  \includegraphics[width=1\linewidth]{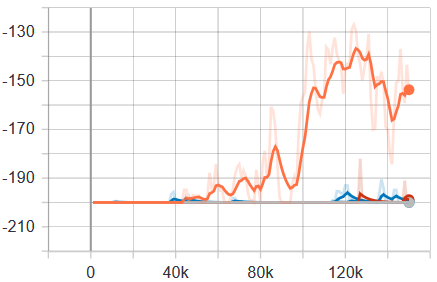}
  \caption{MountainCar-v0 (150k steps)}
\end{subfigure}%
\begin{subfigure}{.233\textwidth}
  \centering
  \includegraphics[width=1\linewidth]{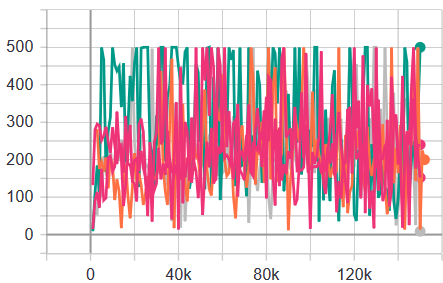}
  \caption{CartPole-v1 (150k steps)}
\end{subfigure}%
\caption{Average return during training of the final model variation with different model dimensions on control environments}
\label{fig: base_performance2}
\end{figure}

\begin{figure} 
\centering
\begin{subfigure}{.233\textwidth}
  \centering
  \includegraphics[width=1\linewidth]{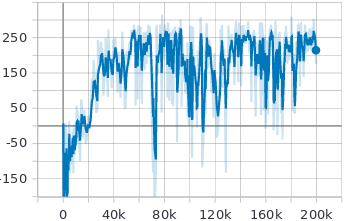}
  \caption{LunarLander-v2 (200k steps)}
\end{subfigure}%
\begin{subfigure}{.233\textwidth}
  \centering
  \includegraphics[width=1\linewidth]{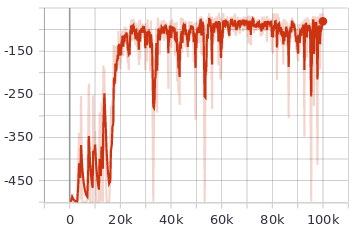}
  \caption{Acrobot-v1 (100k steps)}
\end{subfigure}
\begin{subfigure}{.233\textwidth}
  \centering
  \includegraphics[width=1\linewidth]{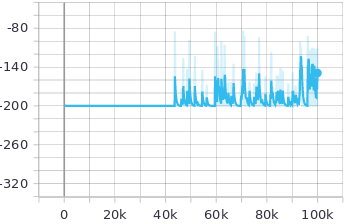}
  \caption{MountainCar-v0 (100k steps)}
\end{subfigure}%
\begin{subfigure}{.233\textwidth}
  \centering
  \includegraphics[width=1\linewidth]{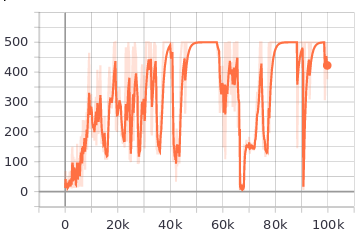}
  \caption{CartPole-v1 (100k steps)}
\end{subfigure}%
\caption{Average return during training of a HP optimized classic Q-learning agent}
\label{fig: base_performance_zoo}
\end{figure}

To evaluate the performance of the method variation specified in table \ref{table:3}, its average return during training was compared to the average return during training of an optimized classic Q-learning agent in four Environments (CartPole-v1, Acrobot-v1, MountainCar-v0, and LunarLander-v2). The method was extracted from  Rl-zoo baselines \cite{rl-zoo}, a collection of Hyperparameter optimized methods. Additionally, the final method variation was tested with different values for history length, model dimensions, and the number of layers (Appendix C) to secure that it is stable and performs consistently when scaled up or down. By examining Figure \ref{fig: base_performance2}, it is visible that the performance of the proposed model is consistent over different model sizes. While the method variation is consistent for CartPole-v1, only one variation is displayed to keep the visibility of the results.
Furthermore, when comparing Figure \ref{fig: base_performance2} with Figure \ref{fig: base_performance_zoo}, it is visible that the average return during training of these different approaches is largely comparable. Noteworthy, the proposed method variation seems to have problems to converge for CartPole-v1. The frequency in which the maximum reward is achieved seems to however increase over time. 

\subsection{Performance in ATARI environments}

\begin{figure} [ht] 
\centering
\begin{subfigure}{.233\textwidth}
  \centering
  \includegraphics[width=1\linewidth]{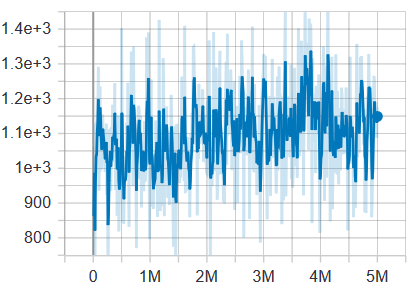}
  \caption{Atari Asteroids}
\end{subfigure}%
\begin{subfigure}{.233\textwidth}
  \centering
  \includegraphics[width=1\linewidth]{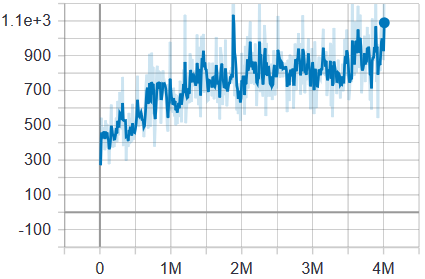}
  \caption{Atari MsPacman 1 }
\end{subfigure}
\begin{subfigure}{.233\textwidth}
  \centering
  \includegraphics[width=1\linewidth]{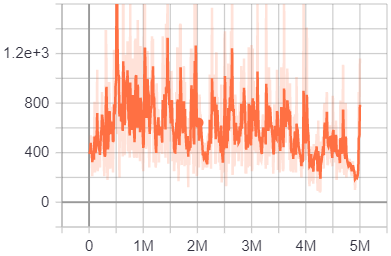}
  \caption{Atari MsPacman 2 }
\end{subfigure}
\caption{Average return during training on Atari}
\label{fig: Atari_av}
\end{figure}

 Additionally to the control environments, the proposed method variation was trained in two environments ("MsPacman" and "Asteroids") of the popular ATARI benchmark. Two parameters that are not included in \ref{table:Final} were scaled up from their initial values to match the complexity of the new environment and to keep them in reasonable ranges. The initial collect steps were increased from 1000 to 5000. Additionally, the Replay Buffer size was increased from 100k to 200k. For both environments, the RAM state which consists out of 128 pixels was used as the state vector for training. The proposed method was trained on MsPacman for 4 and 5 million timesteps and on "Asteroids" on 5 million timesteps. The best average return during training was compared to the reported results from \cite{mnih2015human} (classic DQN performance), and \cite{hausknecht2015deep} (RQN performance). Both studies trained their methods for 10 million timesteps before reporting their final average returns. 

When tracking the method state with the best average return during training for the "Asteroids" environment, the proposed method variation performs quite well. After only 5 million timesteps, the Q-learning agent achieved a higher average return than the reported RQN-based and DQN-based methods. However, during training, the model performance fluctuates strongly which makes the final Q-learning agent perform quite bad. For the "MsPacman" environment, no superior performance can be reported. Additionally, the training behavior does vary significantly. The same TBQN-based Q-learning agent was trained twice. The first try (\ref{fig: Atari_av}b) shows consistent learning over the whole training period. The second one (\ref{fig: Atari_av}c) shows no increase in performance over the whole training. 
While TBQN based methods can perform in ATARI environments, it is still a challenging task. More experiments must be conducted to form a final conclusion. It is also suspected that conducting the parameter search in control environments, might have had a negative effect on the performance of TBQNs in ATARI environments. We are positive, that this challenge can however be overcome by committing more computational resources in the future.

\begin{table}[ht]
\centering
\begin{tabular}{ |p{1.6cm}|p{2.6cm}|p{2.6cm}| }
 \hline
 Methods & Asteroids & MsPacman \\
 \hline
  \hline
 DQN$^2$ & 1629 +/- 542 & 2311 +/- 525 \\
  DQN$^1$ & 1070+/-345 & 2363 +/-735 \\
   RQN$^1$ & 1020 +/-312 & 2048+/-653 \\
   TBQN & 1813+/- 396 &  1555+/-696 \\

 \hline
\end{tabular}

\caption{Reported average returns of different methods on Atari. 1 = \cite{hausknecht2015deep}, 2 = \cite{mnih2015human}}
\label{table:Atari_performance}
\end{table}

\section{Conclusion}

During this work, the interaction of Transformer architectures and Deep Q-learning was evaluated. The goal of this work was to craft a new RL method based on the combination of Deep Q-learning and Transformer-based models which was successful. Through an extensive method variation search, a Transformer-based Deep Q-Learning method was constructed which leverages developments around Transformers as well as Q-learning. The proposed model can match the performance of an optimized classic Q-learning agent on control environments while showing potential on selected Atari environments. Despite these successes, the testing of the proposed final method variation on more environments and especially environments that require a deep understanding of past states is still essential to form a final conclusion. The results of this work are complementary to \cite{parisotto2019stabilizing} and another step to a better understanding of Transformer architectures in RL. This work defies past results that neglect Transformer architectures in RL and shows that they can perform when handled carefully. While the proposed method is connected to the one that was used in \cite{parisotto2019stabilizing}, it represents a different version of a Transformer-based RL method that can be deployed, tuned, and tested more easily. To further encourage this, the code base of this research can be accessed under \cite{GS}. It is hoped that this work can help to support new studies on the topic of Transformers in RL and leverage them to RL mainstream.

\section{Appendix}

\subsection{A. Model specifications}

Throughout this work, the TBQN dimensions specified in Table \ref{table:4} were used. 

\begin{table}[ht]
\centering
\begin{tabular}{ |p{3cm}||p{1.8cm}|p{1.8cm}| }
 \hline
 Specification & Control & Atari \\
 \hline 
   \hline
 History horizon   & 5 steps & 4 steps   \\
 Encoding Dimension   & 64 & 64   \\
 Number of Layers   & 3 & 2   \\
 Dff Dimension  & 256 & 256  \\
 \hline
\end{tabular}
\caption{TBQN dimensions throughout this work}
\label{table:4}
\end{table}

\subsection{B. Study specifications}

Table \ref{table:5} and Table \ref{table:6} hold additional information concerning the studies that were conducted to arrive at a final method variation.
\begin{table}[ht]
\centering
\begin{tabular}{ |p{4cm}||p{3cm}| }
 \hline
 Specification & Value \\
 \hline
   \hline
 Number of evaluated search space samples   & 30   \\
 Number of environments & 3 \\
 Runs per sample & 2 \\
 Training steps  & 15k  \\
 \hline
\end{tabular}
\caption{Additional information for study 1}
\label{table:5}
\end{table}

 \begin{table}[!ht]
\centering
\begin{tabular}{ |p{4cm}||p{3cm}| }
 \hline
 Specification & Value\\
 \hline
 \hline
 Remaining search  space samples   & 24   \\
 Number of environments & 3 \\
 Runs per sample & 2 \\
 Training steps  & 150k / 75k  \\
 \hline
\end{tabular}
\caption{Additional information for study 2}
\label{table:6}
\end{table}

\subsection{C. Model dimension variants}
During the evaluation of the final TBQN variation, the model dimensions were altered to test for stability when scaling TBQNs up or down. The following variations were tested: 

\begin{itemize}
    \item History horizon: 5, Dimensions: 64/256, Layers: 3
    \item History horizon: 5, Dimensions: 64/256, Layers: 6
    \item History horizon: 3, Dimensions: 64/256, Layers: 3
    \item History horizon: 7, Dimensions: 64/256, Layers: 3
    \item History horizon: 5, Dimensions: 128/512, Layers: 3
\end{itemize}

\subsection{D. Additional comments}

All experiments and studies were conducted on a single GPU (Nvidia1080Ti). Specific parameters that are not explicitly defined are set to the default values of TensorFlow \cite{tensorflow2015-whitepaper} or are defined in the experiment scripts available at \cite{GS}

\bibliography{main.bib}

\begin{thebibliography}{22}
\providecommand{\natexlab}[1]{#1}
\providecommand{\url}[1]{\texttt{#1}}
\providecommand{\urlprefix}{URL }
\expandafter\ifx\csname urlstyle\endcsname\relax
  \providecommand{\doi}[1]{doi:\discretionary{}{}{}#1}\else
  \providecommand{\doi}{doi:\discretionary{}{}{}\begingroup
  \urlstyle{rm}\Url}\fi

\bibitem[{Abadi et~al.(2015)Abadi, Agarwal, Barham, Brevdo, Chen, Citro,
  Corrado, Davis, Dean, Devin, Ghemawat, Goodfellow, Harp, Irving, Isard, Jia,
  Jozefowicz, Kaiser, Kudlur, Levenberg, Man\'{e}, Monga, Moore, Murray, Olah,
  Schuster, Shlens, Steiner, Sutskever, Talwar, Tucker, Vanhoucke, Vasudevan,
  Vi\'{e}gas, Vinyals, Warden, Wattenberg, Wicke, Yu, and
  Zheng}]{tensorflow2015-whitepaper}
Abadi, M.; Agarwal, A.; Barham, P.; Brevdo, E.; Chen, Z.; Citro, C.; Corrado,
  G.~S.; Davis, A.; Dean, J.; Devin, M.; Ghemawat, S.; Goodfellow, I.; Harp,
  A.; Irving, G.; Isard, M.; Jia, Y.; Jozefowicz, R.; Kaiser, L.; Kudlur, M.;
  Levenberg, J.; Man\'{e}, D.; Monga, R.; Moore, S.; Murray, D.; Olah, C.;
  Schuster, M.; Shlens, J.; Steiner, B.; Sutskever, I.; Talwar, K.; Tucker, P.;
  Vanhoucke, V.; Vasudevan, V.; Vi\'{e}gas, F.; Vinyals, O.; Warden, P.;
  Wattenberg, M.; Wicke, M.; Yu, Y.; and Zheng, X. 2015.
\newblock {TensorFlow}: Large-Scale Machine Learning on Heterogeneous Systems.
\newblock \urlprefix\url{http://tensorflow.org/}.
\newblock Software available from tensorflow.org.

\bibitem[{Bergstra et~al.(2011)Bergstra, Bardenet, Bengio, and
  K{\'e}gl}]{bergstra2011algorithms}
Bergstra, J.~S.; Bardenet, R.; Bengio, Y.; and K{\'e}gl, B. 2011.
\newblock Algorithms for hyper-parameter optimization.
\newblock In \emph{Advances in neural information processing systems},
  2546--2554.

\bibitem[{Brockman et~al.(2016)Brockman, Cheung, Pettersson, Schneider,
  Schulman, Tang, and Zaremba}]{1606.01540}
Brockman, G.; Cheung, V.; Pettersson, L.; Schneider, J.; Schulman, J.; Tang,
  J.; and Zaremba, W. 2016.
\newblock OpenAI Gym.

\bibitem[{Dai et~al.(2019)Dai, Yang, Yang, Carbonell, Le, and
  Salakhutdinov}]{dai2019transformer}
Dai, Z.; Yang, Z.; Yang, Y.; Carbonell, J.; Le, Q.~V.; and Salakhutdinov, R.
  2019.
\newblock Transformer-xl: Attentive language models beyond a fixed-length
  context.
\newblock \emph{arXiv preprint arXiv:1901.02860} .

\bibitem[{Devlin et~al.(2018)Devlin, Chang, Lee, and
  Toutanova}]{devlin2018bert}
Devlin, J.; Chang, M.-W.; Lee, K.; and Toutanova, K. 2018.
\newblock Bert: Pre-training of deep bidirectional transformers for language
  understanding.
\newblock \emph{arXiv preprint arXiv:1810.04805} .

\bibitem[{Hausknecht and Stone(2015)}]{hausknecht2015deep}
Hausknecht, M.; and Stone, P. 2015.
\newblock Deep recurrent q-learning for partially observable mdps.
\newblock In \emph{2015 AAAI Fall Symposium Series}.

\bibitem[{Iqbal and Sha(2019)}]{iqbal2019actor}
Iqbal, S.; and Sha, F. 2019.
\newblock Actor-attention-critic for multi-agent reinforcement learning.
\newblock In \emph{International Conference on Machine Learning}, 2961--2970.
  PMLR.

\bibitem[{Lillicrap et~al.(2015)Lillicrap, Hunt, Pritzel, Heess, Erez, Tassa,
  Silver, and Wierstra}]{lillicrap2015continuous}
Lillicrap, T.~P.; Hunt, J.~J.; Pritzel, A.; Heess, N.; Erez, T.; Tassa, Y.;
  Silver, D.; and Wierstra, D. 2015.
\newblock Continuous control with deep reinforcement learning.
\newblock \emph{arXiv preprint arXiv:1509.02971} .

\bibitem[{Manchin, Abbasnejad, and van~den
  Hengel(2019)}]{manchin2019reinforcement}
Manchin, A.; Abbasnejad, E.; and van~den Hengel, A. 2019.
\newblock Reinforcement learning with attention that works: A self-supervised
  approach.
\newblock In \emph{International Conference on Neural Information Processing},
  223--230. Springer.

\bibitem[{Mishra et~al.(2017)Mishra, Rohaninejad, Chen, and
  Abbeel}]{mishra2017simple}
Mishra, N.; Rohaninejad, M.; Chen, X.; and Abbeel, P. 2017.
\newblock A simple neural attentive meta-learner.
\newblock \emph{arXiv preprint arXiv:1707.03141} .

\bibitem[{Mnih et~al.(2013)Mnih, Kavukcuoglu, Silver, Graves, Antonoglou,
  Wierstra, and Riedmiller}]{mnih2013playing}
Mnih, V.; Kavukcuoglu, K.; Silver, D.; Graves, A.; Antonoglou, I.; Wierstra,
  D.; and Riedmiller, M. 2013.
\newblock Playing atari with deep reinforcement learning.
\newblock \emph{arXiv preprint arXiv:1312.5602} .

\bibitem[{Mnih et~al.(2015)Mnih, Kavukcuoglu, Silver, Rusu, Veness, Bellemare,
  Graves, Riedmiller, Fidjeland, Ostrovski et~al.}]{mnih2015human}
Mnih, V.; Kavukcuoglu, K.; Silver, D.; Rusu, A.~A.; Veness, J.; Bellemare,
  M.~G.; Graves, A.; Riedmiller, M.; Fidjeland, A.~K.; Ostrovski, G.; et~al.
  2015.
\newblock Human-level control through deep reinforcement learning.
\newblock \emph{Nature} 518(7540): 529--533.

\bibitem[{Oh et~al.(2016)Oh, Chockalingam, Singh, and Lee}]{oh2016control}
Oh, J.; Chockalingam, V.; Singh, S.; and Lee, H. 2016.
\newblock Control of memory, active perception, and action in minecraft.
\newblock \emph{arXiv preprint arXiv:1605.09128} .

\bibitem[{Parisotto et~al.(2019)Parisotto, Song, Rae, Pascanu, Gulcehre,
  Jayakumar, Jaderberg, Kaufman, Clark, Noury
  et~al.}]{parisotto2019stabilizing}
Parisotto, E.; Song, H.~F.; Rae, J.~W.; Pascanu, R.; Gulcehre, C.; Jayakumar,
  S.~M.; Jaderberg, M.; Kaufman, R.~L.; Clark, A.; Noury, S.; et~al. 2019.
\newblock Stabilizing Transformers for Reinforcement Learning.
\newblock \emph{arXiv preprint arXiv:1910.06764} .

\bibitem[{Radford et~al.(2019)Radford, Wu, Child, Luan, Amodei, and
  Sutskever}]{radford2019language}
Radford, A.; Wu, J.; Child, R.; Luan, D.; Amodei, D.; and Sutskever, I. 2019.
\newblock Language models are unsupervised multitask learners.
\newblock \emph{OpenAI Blog} 1(8): 9.

\bibitem[{Raffin(2018)}]{rl-zoo}
Raffin, A. 2018.
\newblock RL Baselines Zoo.
\newblock \urlprefix\url{https://github.com/araffin/rl-baselines-zoo}.

\bibitem[{Song et~al.(2019)Song, Abdolmaleki, Springenberg, Clark, Soyer, Rae,
  Noury, Ahuja, Liu, Tirumala et~al.}]{song2019v}
Song, H.~F.; Abdolmaleki, A.; Springenberg, J.~T.; Clark, A.; Soyer, H.; Rae,
  J.~W.; Noury, S.; Ahuja, A.; Liu, S.; Tirumala, D.; et~al. 2019.
\newblock V-MPO: On-Policy Maximum a Posteriori Policy Optimization for
  Discrete and Continuous Control.
\newblock \emph{arXiv preprint arXiv:1909.12238} .

\bibitem[{Stein(2020)}]{GS}
Stein, G. 2020.
\newblock Gideon-Stein/TBQN.
\newblock \urlprefix\url{https://github.com/Gideon-Stein/TBQN}.

\bibitem[{Upadhyay et~al.(2019)Upadhyay, Shah, Ravikanti, and
  Medhe}]{upadhyay2019transformer}
Upadhyay, U.; Shah, N.; Ravikanti, S.; and Medhe, M. 2019.
\newblock Transformer Based Reinforcement Learning For Games.
\newblock \emph{arXiv preprint arXiv:1912.03918} .

\bibitem[{Vaswani et~al.(2017)Vaswani, Shazeer, Parmar, Uszkoreit, Jones,
  Gomez, Kaiser, and Polosukhin}]{vaswani2017attention}
Vaswani, A.; Shazeer, N.; Parmar, N.; Uszkoreit, J.; Jones, L.; Gomez, A.~N.;
  Kaiser, {\L}.; and Polosukhin, I. 2017.
\newblock Attention is all you need.
\newblock In \emph{Advances in neural information processing systems},
  5998--6008.

\bibitem[{Xiong et~al.(2020)Xiong, Yang, He, Zheng, Zheng, Xing, Zhang, Lan,
  Wang, and Liu}]{xiong2020layer}
Xiong, R.; Yang, Y.; He, D.; Zheng, K.; Zheng, S.; Xing, C.; Zhang, H.; Lan,
  Y.; Wang, L.; and Liu, T.-Y. 2020.
\newblock On layer normalization in the transformer architecture.
\newblock \emph{arXiv preprint arXiv:2002.04745} .

\bibitem[{Zhang, Titov, and Sennrich(2019)}]{zhang2019improving}
Zhang, B.; Titov, I.; and Sennrich, R. 2019.
\newblock Improving deep transformer with depth-scaled initialization and
  merged attention.
\newblock \emph{arXiv preprint arXiv:1908.11365} .

\end{thebibliography}

\end{document}